# Developing a Calibrated Physics-Based Digital Twin for Construction Vehicles

Deniz Karanfil[1,2,3,4], Daniel Lindmark[5,7], Martin Servin[5,6,8], David Torick[2,9], Bahram Ravani[1,2,10]

[1] *Department of Mechanical and Aerospace Engineering, University of California, Davis, CA 95616, USA*

[2] *Advanced Highway Maintenance and Construction Technology (AHMCT) Research Center at the University of California, Davis, CA 95616, USA*

[3] *dkaranfil@ucdavis.edu (Corresponding author)*

[4] *This paper presents research conducted as part of the first author's Ph.D. dissertation entitled Developing Scalable Digital Twins of Construction Vehicles.*

[5] *Algoryx Simulation, Uminova Science Park, Kuratorvägen 2B, 907 36 Umeå, Sweden*

[6] *Department of Physics, Umea University, MIT-huset, byggnad V, UMIT Research Lab, 901 87 Umea, Sweden*

[7] *daniel.lindmark@algoryx.com*

[8] *martin.servin@umu.se*

[9] *datorick@ucdavis.edu*

[10] *bravani@ucdavis.edu*

## Abstract

This paper presents the development of a calibrated digital twin of a wheel loader. A calibrated digital twin integrates a construction vehicle with a high-fidelity digital model allowing for automated diagnostics and optimization of operations as well as pre-planning simulations enhancing automation capabilities. The high-fidelity digital model is a virtual twin of the physical wheel loader. It uses a physics-based multibody dynamic model of the wheel loader in the software AGX Dynamics. Interactions of the wheel loader's bucket while in use in construction can be simulated in the virtual model. Calibration makes this simulation of high-fidelity which can enhance realistic planning for automation of construction operations. In this work, a wheel loader was instrumented with several sensors used to calibrate the digital model. The calibrated digital twin was able to estimate the magnitude of the forces on the bucket base with high accuracy, providing a high-fidelity simulation.

## 1. Introduction

A Digital twin is a computer model of a physical object, a system, or a process that is tightly integrated with its physical counterpart through data exchange, typically via sensors and data acquisition systems [1,2]. This integration provides a high-fidelity virtual environment tailored for a broad range of applications. These include analyzing, diagnosing or observing the behavior of the physical twin (the system) [3,4], optimizing its performance without needing to directly work with the physical counterpart of the system, enhancing safety [5-7], facilitating proactive maintenance [8,9], and numerous other engineering and decision-making tasks. While some digital twins are developed using data-driven approaches relying on information such as visual data and operational parameters, others are based on physics-based simulations [8-11] that model system behavior from its governing equations. One main difference between a digital twin and a conventional computer simulation is that a digital twin is usually integrated into the physical system through sensors and data acquisition systems, which are embedded in the physical system [12]. Physics-based simulations have long been developed for heavy construction equipment [13-16]. In the case of a wheel loader, for example, a good physics-based simulation system should not only include the dynamics of the equipment but also the physics of the interaction of the end-loader bucket with the soil. In this paper we use AGX Dynamics [17,27] for the physics-based computer model of our digital twin. This is an advanced simulation software which uses a non-smooth contacting multibody dynamics modelling paradigm [18,19] for simulating any wheel loader combined with a mechanistic model for the soil dynamics, that combines a volumetric and particle-based representation of soil interaction using Hertz-Mindlin contact theory [20-22] with discrete element method [23,24]. In this work, we integrate this simulation software with a physical wheel loader and perform experimental interactions with different soils to calibrate the parameters of the digital model in making our digital twin a more realistic and high-fidelity virtual environment for a wheel loader. Enabling automated diagnostics, performance optimization, and pre-planning for autonomous operations of construction vehicles necessitates the accurate prediction of key state parameters. Among these, the forces exerted on the bucket constitute a particularly critical variable. Ensuring the accuracy of the model in predicting these forces is essential for leveraging it in automation-related applications such as autonomous task planning, performance optimization, intelligent control and so on [25, 26] In this manner the digital twin will become calibrated with respect to interacting forces in the real operation of the system through offline instrumentation and data collection from the physical system.

The gap between computer simulation to reality for wheel loaders has been recognized in the past and some methods of optimization have been developed (see for example, 27). Furthermore, as digital twins become increasingly integrated across diverse industries such as manufacturing [28], shipbuilding [29], and transportation [30], there has also been growing interest in their application within the construction machinery sector. Previous studies have documented several implementations of digital twins in this domain [2,31]. For example, Hitachi Construction Machinery has developed a digital twin that focuses on management of construction sites and monitoring machinery operation and construction progress [32]; Hussain et al. present a digital-twin that predicts degraded lifting capacity in tower cranes [33]; whereas Baur & Teutsch developed a physics-based excavator digital twin for hinge-bearing wear prediction [34]. With the growing interest in integrating autonomous features into wheel loaders [35], due to cost and efficiency considerations [36-38], incorporating physics-based digital twins into the wheel loader industry could greatly enhance the integration of such autonomous capabilities. However, the existing body of work in regard to physics-based wheel loader digital twins is largely limited to conceptual studies [39-41]. A recent multibody modelling-based twin for a tractor front-loader by Tuerlinckx et al. focuses on payload-driven stability visualization but does not model or calibrate for bucket–soil interaction forces [42]. Meanwhile, Wu et al [25] have employed a data-driven approaches: first with a Convolutional Neural Network (CNN)-Transformer transfer-learning twin to predict loading resistance; and, more recently, a deep-incremental-learning model that reconstructs and continuously updates a surrogate from a DEM–MBD simulation [26] There remains a gap in the literature for implementation of physics-based digital twins calibrated with real world data and validated with in-situ force and pressure measurements for the hydraulics and the bucket, which is addressed in this paper. In other domains, digital twin systems have been developed that integrate the computer models with the physical system (see, for example [43]) or with a library of components [44].

In general, physics-based digital twins, when calibrated with actual real world interaction data, can provide a basis for integrating autonomous features to construction vehicles [45,46]. They can also be used for analysis to improve the design of the system in the heavy machinery industry. For instance, the specific framework presented in this study has been utilized to develop excavation force estimation [47] and optimal hybrid powertrain design and control [48].

## 2. Methods

### 2.1. Modeling

A good digital twin of a wheel loader first requires a physics-based digital model which starts with representing the form and the function of the device. The form of the device is captured in a CAD model and the function is represented with a multibody kinematics and dynamics model. In addition, there is a need for a physics-based model of the interaction of the end-bucket with the environment as it is used in earthmoving operations in a construction setting. In this work, the CAD model of the wheel loader was developed based on the measurements of the dimensions of the bucket, the hydraulic actuators, the joints, and arms of the front mechanism, and other structural components of a physical wheel loader. These parts of the wheel loader constitute its articulated mechanism, and the generated CAD model of the wheel loader is depicted in Figure 1, including the labeling of the CAD models of the articulated mechanism's components. The CAD model of the wheel loader was imported into the AGX Dynamics simulation framework and assigned mass properties of the components and their associated kinematic topology. The mass properties used are listed in Table 1 and were generated based on the volumes calculated by the software based on the measurements taken from the physical wheel loader itself. The simulation software is then used to capture the function of the articulated mechanism of the wheel loader using a multibody dynamics model of its operation.

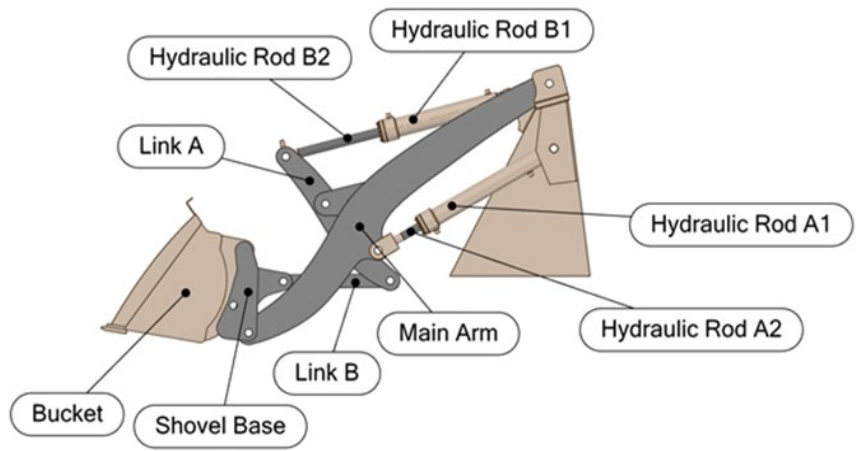

Figure 1. The components of the end-loader CAD model.

The AGX digital twin model's interaction of the end-bucket with the environment (namely soil in earth removing operation) uses a solid mechanics type contact model based on a multiscale model for deformable terrain. The kinematic model of the articulated mechanism is used as constraints in the multibody dynamics model generated within AGX Dynamics. The digital twin presented in this paper is only for the articulated mechanism of the wheel loader and the movement of the actual vehicle is simulated by a main body positioning actuator. This is sufficient for simulating loading unloading operation of a wheel loader. Integration of the movement of the vehicle can be an area of future research. The movement of all the pieces of the articulated mechanism of the wheel loader are indicated by arrows in Figure 2 where the two main cylinders are responsible for controlling the position and orientation of the bucket.

Table 1. List of the assigned masses for each component of the end-loader CAD model.

| Part | Mass (kg) |
|---|---|
| Bucket | 205.8 |
| Bucket Base | 84.8 |
| Link B | 8.99 |
| Link A | 33.9 |
| Main Arm | 294.6 |
| Hydraulic Rod A1 | 20.3 |
| Hydraulic Rod A2 | 13.1 |
| Hydraulic Rod B1 | 19.6 |
| Hydraulic Rod B2 | 14.7 |

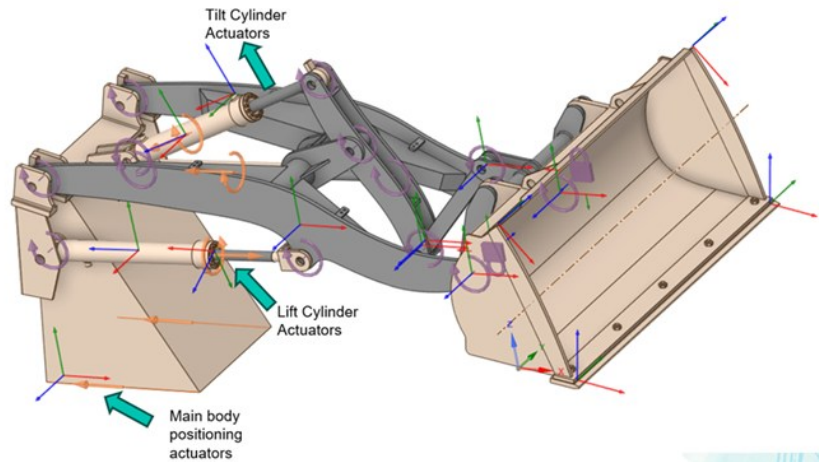

Figure 2. Movement of all the elements of the articulated mechanism of the wheel loader.

### 2.2. Kinematics of the Wheel Loader and Its Mapping Between the Physical Twin and the Simulator

Both the physical and digital twins are controlled with the hydraulic cylinders responsible for lifting the end loader mechanism and tilting the bucket. Kinematic analysis of the end loader mechanism can be utilized to map the height and orientation of the bucket to the positions of the hydraulic cylinders. The end loader mechanism, which has two degrees of freedom [49], each controlled by the hydraulic cylinders. The two symmetrically positioned hydraulic cylinders control the height of the bucket of the end loader mechanism and the single top cylinder controls the orientation of the bucket through a linkage arrangement.

Bucket orientation ($\theta_4$) and the vertical distance between the ground and the bucket blade ($y_{P8}$) are mapped to the extension lengths of the hydraulic cylinders through the inverse kinematics.

$$\begin{pmatrix} \theta_4 \\ y_{P8} \end{pmatrix} \xrightarrow{mapping\ through\ the\ kinematic\ model} \begin{pmatrix} s_1 \\ s_2 \end{pmatrix}$$

### 2.3. Instrumentation of the Physical Twin

The physical wheel loader whose digital twin is presented in this paper is shown in Figure 3. The wheel loader is instrumented with several sensors and the data obtained from them is used to calibrate its digital twin. The first group of sensors integrated into the physical wheel loader consists of an encoder on the rear wheels of the wheel loader to provide dead reckoning data for the vehicle movement and pressure sensors in the two hydraulic cylinders. The pressures in these hydraulic cylinders provide data on the forces encountered in lifting the bucket and those responsible for tilting the bucket. The latter is done by the middle cylinder and the former by the side cylinders.

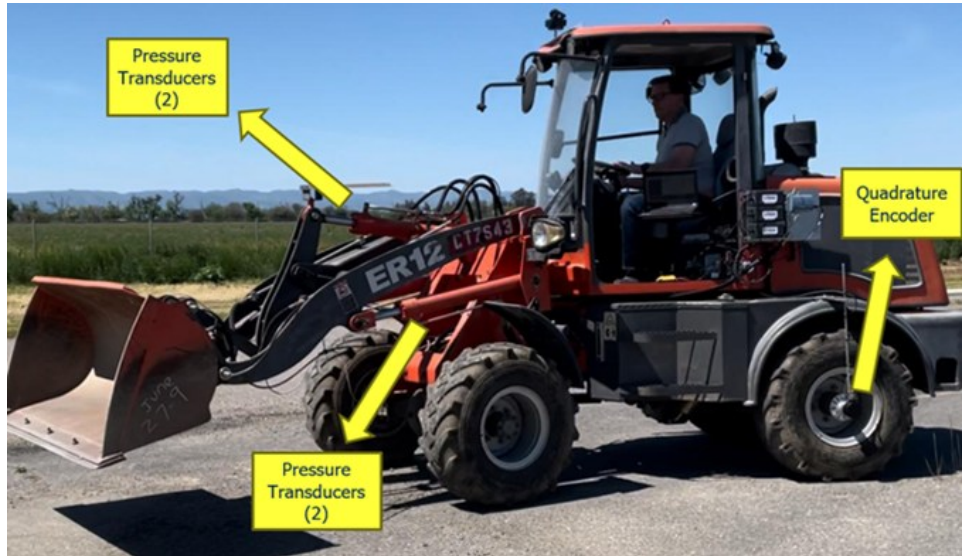

Figure 3. The physical twin.

The second group of sensors integrated into the physical twin are related to the bucket and intended to evaluate the forces of interactions of the bucket with the environment during its operation. The simplest part of this group is an inclinometer which is used to measure the orientation of the bucket. The height of the bucket is measured physically, and these two parameters are used in the inverse kinematic solution to determine the extensions of the cylinders. The next set of sensors were more complicated and required modifications to the hinges connecting the bucket to the articulated mechanism. The three hinge pins connecting the bucket to the articulated mechanism are shown in Figure 4. These hinge pins were removed and remanufactured to include load pins to measure forces experienced at each hinge. Figure 5 shows the remanufactured hinge pin where the load pin is shown to be integrated inside the hinge pin. The load pins used are dual axis load sensors using strain gauges for measuring forces in two directions.

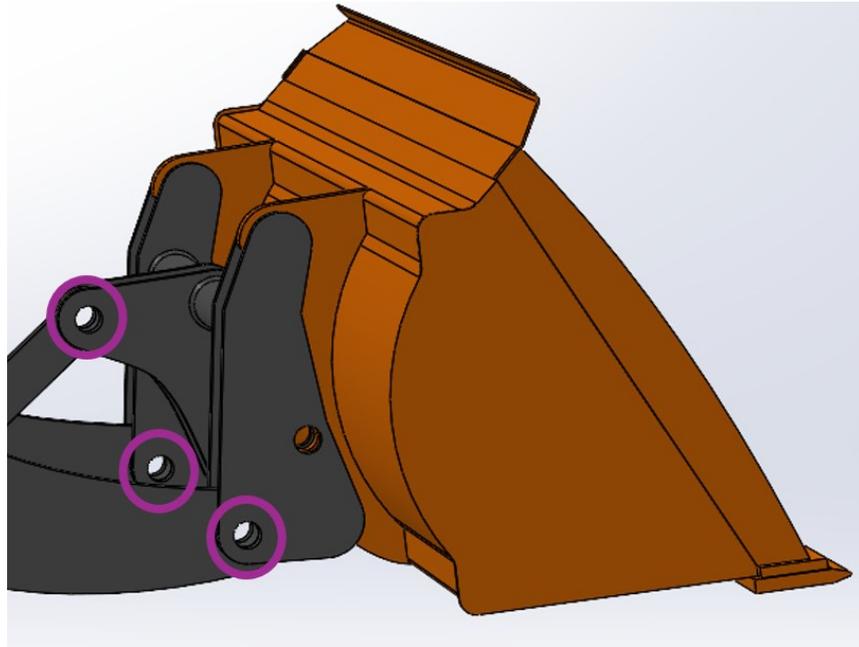

Figure 4. The three hinge pins that were modified.

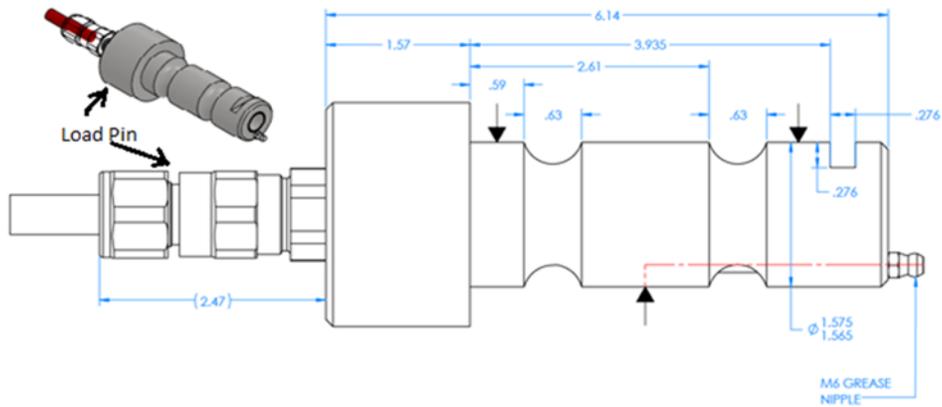

Figure 5. A remanufactured hinge pin with the load pin connected at its end.

The exact locations of this second group of sensors on the physical wheel loader are depicted in Figure 6. The additional instrumentation consisting of data acquisition systems and strain gauge amplifies and circuitry are depicted in Figure 7.

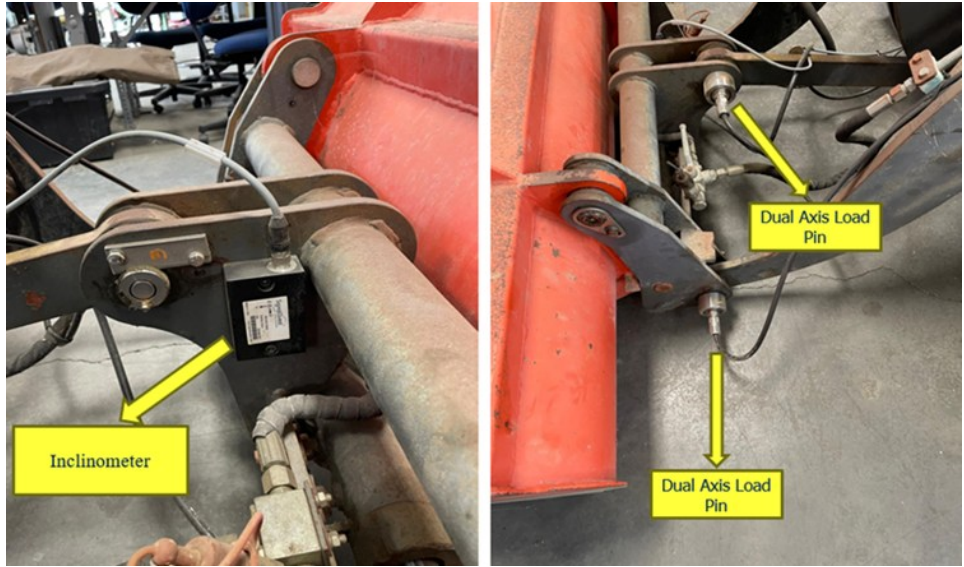

Figure 6. Location of the second group of sensors.

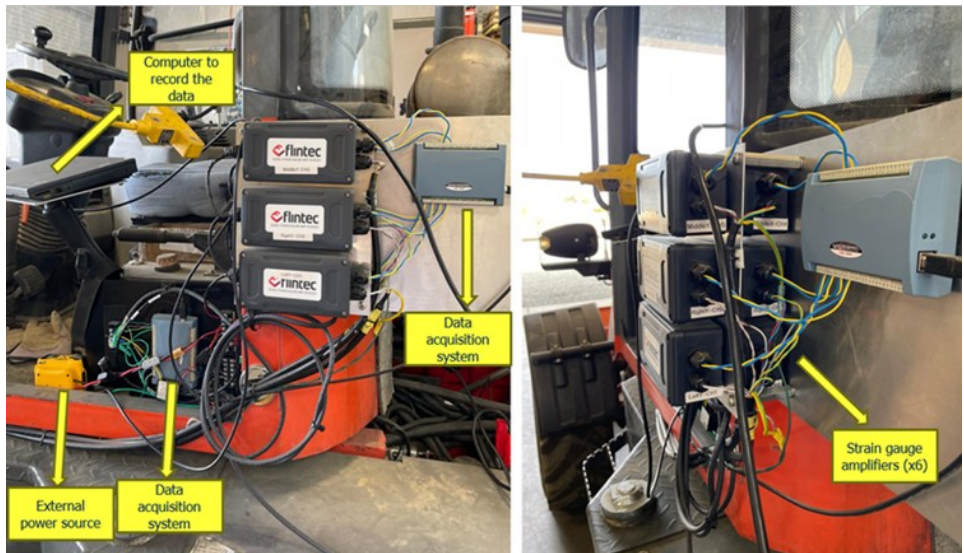

Figure 7. Data collection instrumentation.

2.4. Converting the Load Pin Readings to Forces on The Wheel Loader Bucket

Figure 8 depicts the free-body diagram of the wheel loader's bucket. The mechanism's main arm and link A exert forces on the bucket through locations where pins are mounted. These forces are denoted as $F_{sp_x}$, $F_{sp_y}$, $F_{mp_x}$ and $F_{mp_y}$. Additionally, the bucket has its own weight and the forces due to soil interactions are represented as $F_{s_y}$ and $F_{s_x}$. From the free-body diagram:

$$F_{mp_x} + F_{sp_x} - F_{s_x} = ma_x \quad (1)$$
$$F_{mp_y} + F_{sp_y} + F_{s_y} - W = ma_y \quad (2)$$

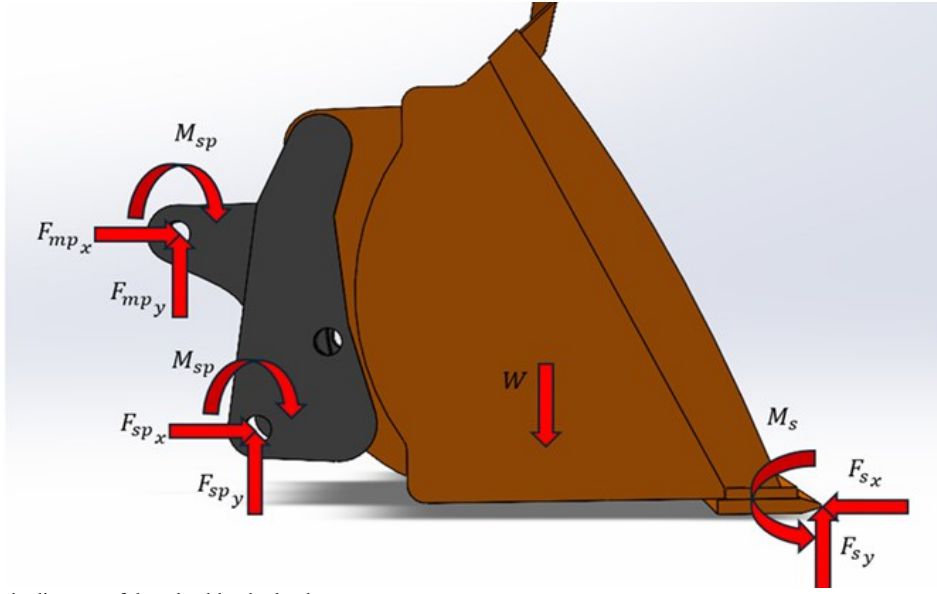

Figure 8. Free body diagram of the wheel loader bucket.

Equations (1) and (2) are simplified for static equilibrium as follows:

$$F_{mp_x} + F_{sp_x} - F_{s_x} = 0 \quad (3)$$

$$F_{mp_y} + F_{sp_y} + F_{s_y} - W = 0 \quad (4)$$

Since the motion of the bucket is planar, the only effective forces are the ones acting along the x- and y- axes.

The pins that are modified experience the forces at the joint. Figure 9 includes hatched areas to indicate where the pin and the components of the mechanism are in direct contact. The design of the pins ensures that every force that is exerted on the bucket by the other components of the mechanism are applied radially via the pins. Figure 10 illustrates the distribution of the loads across the contact areas, where $q_{lA}(s)$ and $q_{bb}(s)$ denote the load distributions resulting from contact with link A and the bucket base, respectively. Meanwhile, $V_1$ and $V_2$ represent the shear forces acting at the grooves where strain gauges are integrated. The strains recorded by the gauges are exclusively induced by the shear stresses, as they are positioned at the neutral axis locations. Moreover, their alignment ensures that strain components can be effectively measured. $V_1$ and $V_2$ can be expressed as follows:

$$V_1 = \int_0^{s_1} q_{bb}(s)ds \quad (5)$$

$$V_2 = \int_{l-s_2}^{l} q_{lA}(s)ds \quad (6)$$

Utilizing $V_1$ and $V_2$, the resultant load that the bucket base and the link exert on the pin is determined.

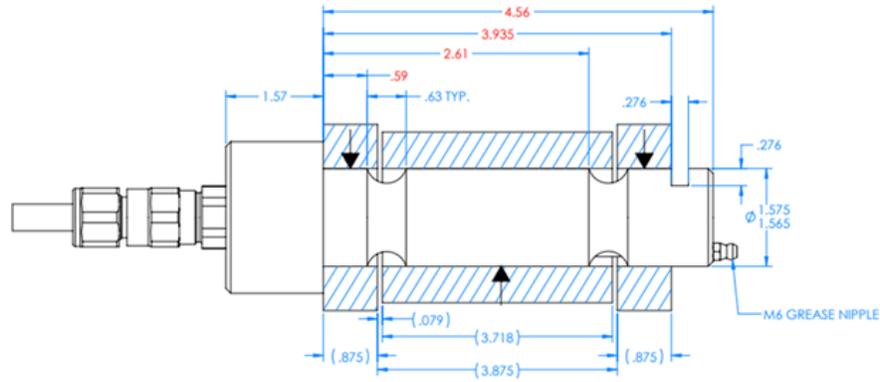

Figure 9. Technical drawing of the load pin.

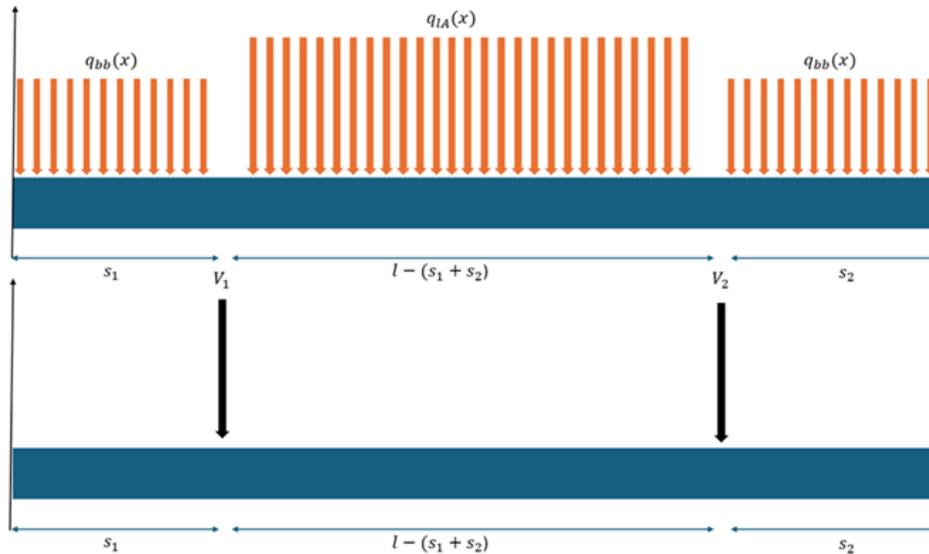

Figure 10. Distribution of the loads applied to the load pin by the bucket base and the link A, and the shear forces corresponding to the cross sections where the strain gauges are inserted.

2.5. Calibration

The calibration was performed in two steps. In the first step, several template trajectories were simulated in the digital model of the physical wheel loader making sure that the kinematics and the movement of the articulated mechanism and the orientation of the bucket matched the movements of the physical wheel loader, its articulated mechanisms, and its end bucket. A side-by-side depiction of the physical and the digital twins is depicted in Figure 11. Data from the integrated sensors [52] and those extracted from the digital simulation related to the movement of the base and the orientation of the bucket are depicted in Figures 12 and 13 respectively. It is clear from these two figures that a good match was observed in the experiment versus the digital simulation in terms of the movements and the orientation of the end bucket.

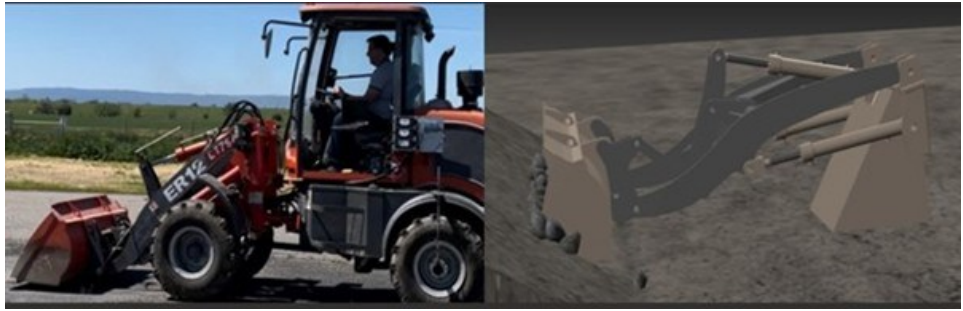

Figure 11. A side-by-side depiction of the physical and the digital twins.

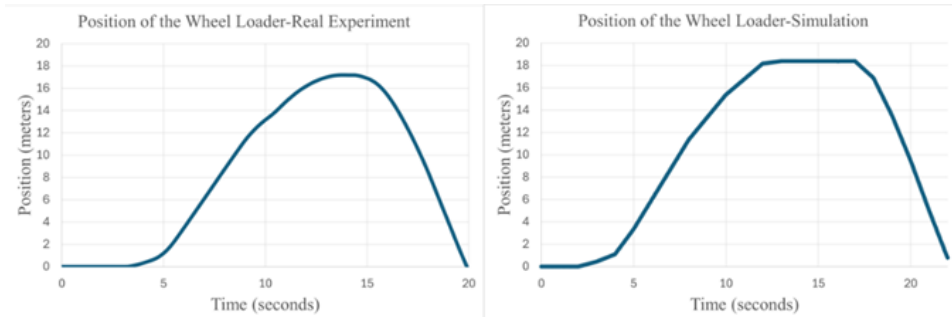

Figure 12. Data indicating movement of the base of the articulated mechanism of the wheel loader in the digital twin versus movement of the vehicle in the physical twin.

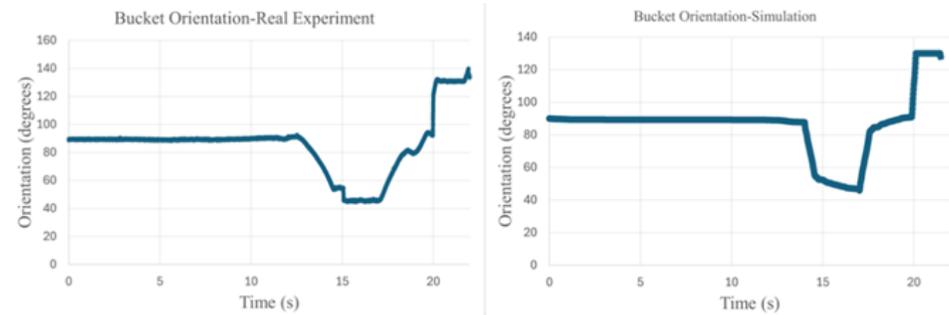

Figure 13. Data indicating the orientation of the end-bucket in the physical and the digital twins.

The second step in calibration involved matching the dynamic forces experienced by the end-bucket between the digital simulation and physical experimentation. This was an iterative process involving incremental changes in the soil parameters used in the digital twin by trial and error until an accurate matching was observed between the digital data and the experimental data. The soil parameters and their updates after calibration are listed in Table 2.

Table 2. Terrain parameter values before and after the calibration process.

| Terrain Parameter | Value Before Calibration | Value After Calibration |
| --- | --- | --- |
| Young's Modulus (MPa) | 1.0 | 20 |
| Friction Coefficient ($\mu_t$) | 0.67 | 0.68 |
| Coefficient of Restitution ($e$) | 0.25 | 0.25 |
| Soil particle Size (m) | 0.06 | 0.06 |
| Rolling Resistance Coefficient ($\mu_r$) | 0.1 | 0.3 |

## 3. Results

Once the calibration was completed data was collected in real experiments using the physical twin and was matched with data extracted from the digital twin. The side-by-side comparison of the lift and tilt cylinder data are shown in Figures 14 and 15 respectively.

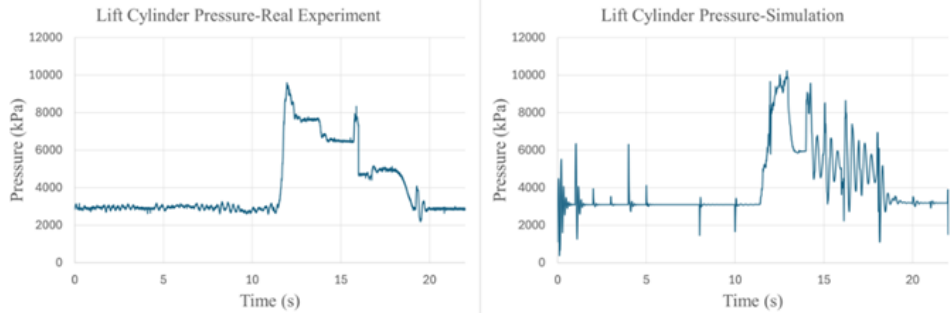

Figure 14. Comparison of the lift cylinder pressure between the physical measurements and the data from the digital twin.

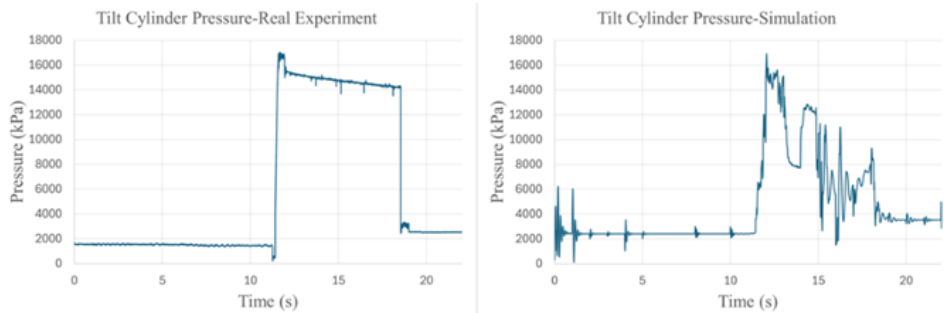

Figure 15. Comparison of the tilt cylinder pressure between the physical measurements and the data from the digital twin.

The data from the tilt cylinder [52] as depicted in Figure 15 is much noisier as compared to the data from the lift cylinder. This is most probably due to additional vibration which occurred during tilting of the bucket as compared to lifting it. Plots of the forces on the bucket without calibration and after calibration during bucket interaction with the soil as compared to the experimental data are shown in Figure 16. The results indicate that physics-based modeling using Multi Body Dynamics accurately captures the physics of the end loader mechanism but lacks accuracy in representing its interaction with the soil. Incorporating calibration enables the model to capture soil interaction physics more effectively, highlighting the essential role of calibration. It is clear from these plots that the calibrated digital twin provides a much better representation of the real system and its operation. The error in the magnitude of the force on the bucket during soil interaction is quantified in Table 3. It is clear from the data in this table that the average error for the force on the end bucket decreased from approximately 74% to approximately only 10%. Furthermore, the error for the maximum force experienced by the end bucket reduced from 118% to approximately 5% after calibration.

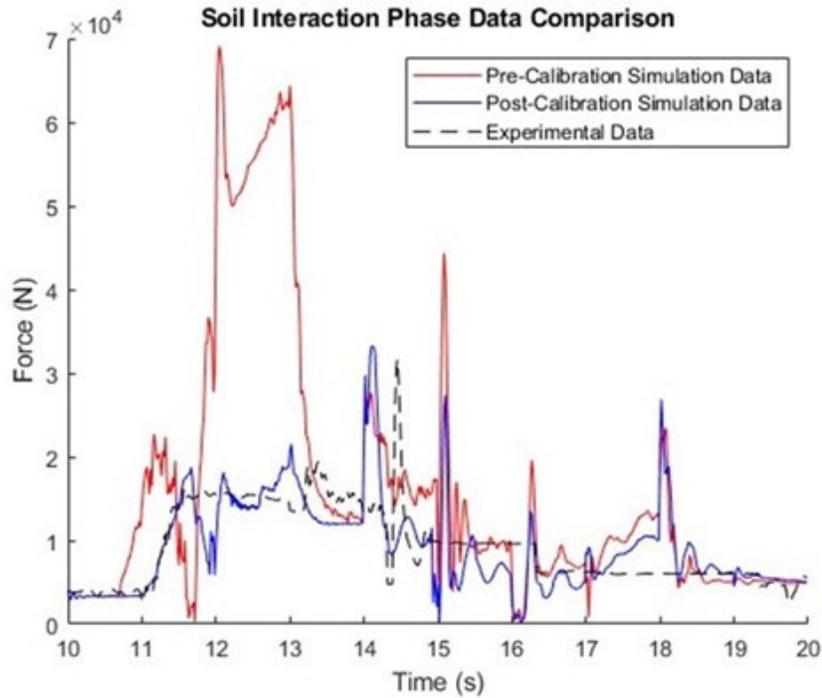

Figure 16. Comparison of forces on the end bucket pre and post calibration with experimental data.

Table 3. Average errors in percentage for the force on the end bucket between pre-calibration and post calibration of the digital twin.

|  | Error for the maximum force (%) | Average error for the force (%) |
| --- | --- | --- |
| Pre-Calibration | 117.96 | 73.62 |
| Post-Calibration | 4.90 | 10.1 |

### 3.1. Conformation of the Calibration with Additional Testing Involving Different Configurations

Figures 17 and 18 compare the tests conducted with the physical and digital twins. These additional tests encompass different configurations of the bucket's height and orientation at various stages of the operation cycle, which facilitates the validation of the calibration process and the calibrated simulator. The same calibrated terrain parameter values from Table 2 were used in the simulator shown in Figures 17 and 18. When evaluating the maximum force throughout the cycle, these simulations yielded errors only ranging between 3.29 to 14.46% for the peak force which demonstrates strong and consistent improvements in accuracy across a variety of operation cycles.

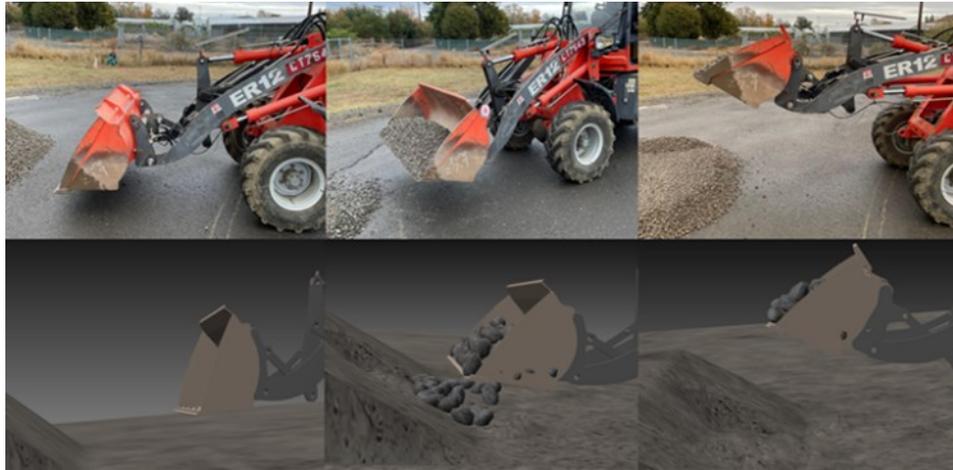

Figure 17. Stages of one of the configurations used for the calibration tests with the operation of the physical wheel loader shown on the upper side and the same operation carried out in the form of multibody simulation shown on the lower side.

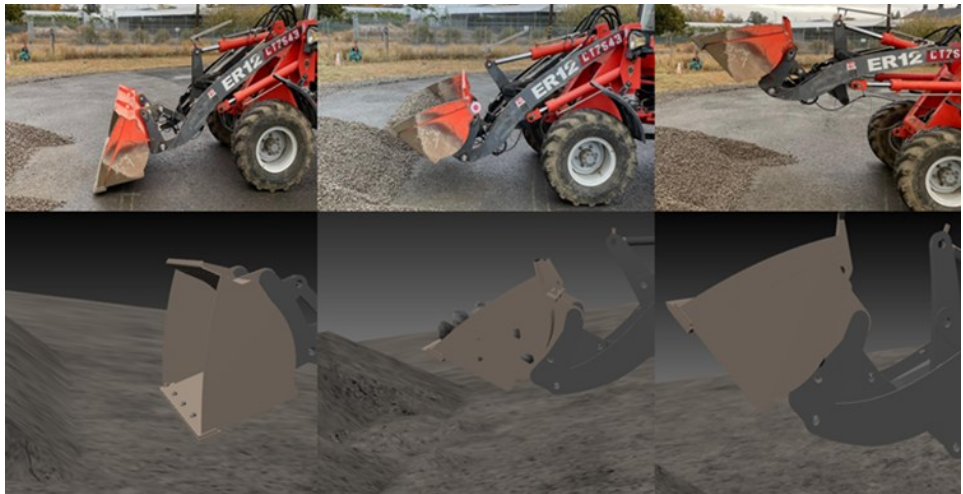

Figure 18. Test carried out with a different configuration and replicated on the simulator.

## 4. Discussion

This paper presented the development of a high-fidelity, calibrated digital twin of a construction vehicle interacting with the environment in earth moving operations. An experimental setup was developed consisting of a digital model based on the well-known multibody simulation software AGX Dynamics and a physical system consisting of an actual wheel loader used for experimentation. The setup also included a set of sensors including pressure transducers, custom made and fabricated dual axis load pins, an inclinometer, and a quadrature encoder. Highly detailed synchronized data for critical parameters such as bucket orientation, loader position, forces on the bucket base and the hydraulic cylinder pressures were collected and used for calibration of the digital twin. The results show that the errors in evaluating interaction forces on the end device of the system can be significantly reduced when the digital twin is calibrated, resulting in a more reliable virtual environment that supports a wide range of automation tasks in construction processes. The digital twin can be effectively utilized for simulation-driven pre-planning of earthmoving operations, the design and validation of intelligent control algorithms, and real-time or offline performance optimization and diagnostics of wheel loaders. By accurately predicting dynamic behaviors and forces during interaction of the vehicle with the environment, the model enables safer, more efficient, and cost-effective construction workflows. Furthermore, the physics-based virtual environment can be utilized as a testbed to evaluate control algorithms or operation sequences which reduce reliance on trial and error and support further automation in heavy construction equipment operations. Although ideally an integrated digital twin needs to have a real time connection to its physical twin, it is shown in this paper that a simpler offline integration can be used for calibration which can significantly increase the accuracy of the digital twin and therefore expand its domain of applications.

**Authors' contributions**

Deniz Karanfil: Writing – original draft, Conceptualization, Visualization, Validation, Methodology, Investigation, Formal analysis, Resources, Data curation, Software. Daniel Lindmark: Resources, Software, Validation, Methodology. Martin Servin: Resources, Software, Validation, Methodology. David Torick: Investigation, Validation, Resources. Bahram Ravani: Writing – review & editing, Project administration, Supervision, Funding acquisition, Conceptualization, Validation, Methodology.

**Disclosure of interest**

The authors report there are no competing interests to declare.

**Funding**

This work was supported in part by funding from Komatsu of Japan. Their support of this work is greatly appreciated.

**Data availability statement**

Supplementary material: The calibration dataset for the digital twin is available at [52]. The supplementary demonstration video of the digital twin is available at [53]. Additional supplementary material related to this study would be made available from the corresponding author upon reasonable request.

**Appendix A. Solution of the Inverse Kinematics**

Figures A.1 to A.4 depict the established kinematic model, whereas Table A.1 lists the values of important parameters.

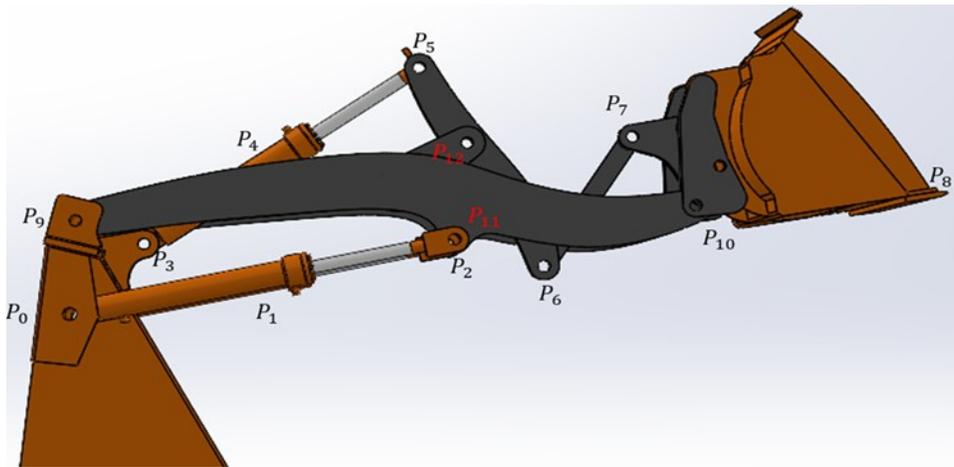

Figure A.1. The points between the linkages of the kinematic model shown on the CAD model of the end loader mechanism.

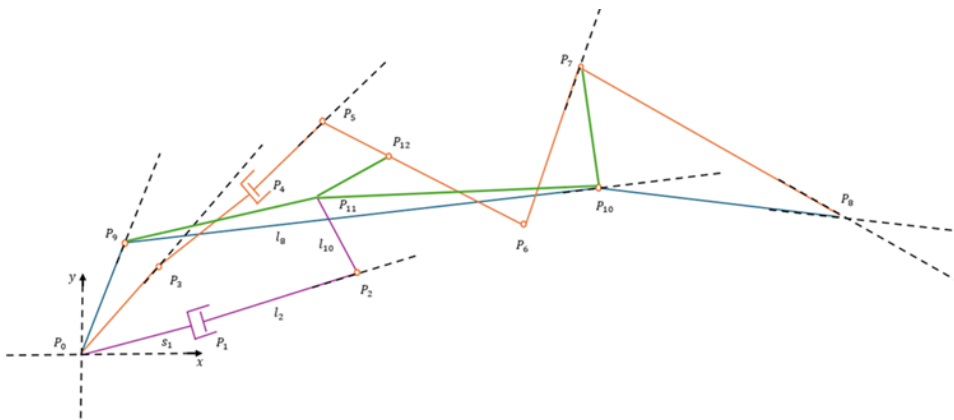

Figure A.2. The points between the linkages of the kinematic model of the end loader mechanism.

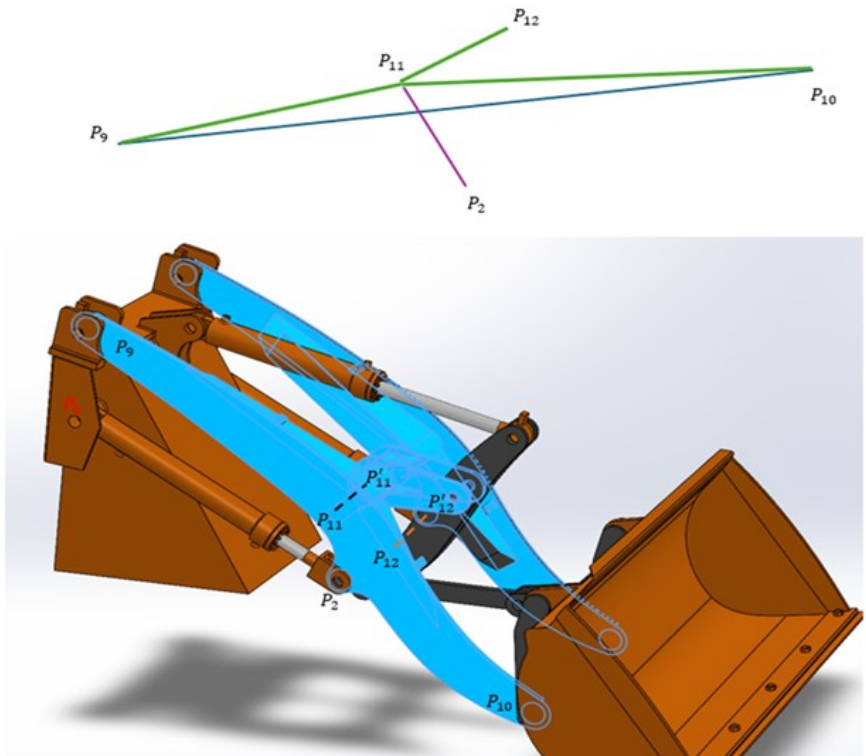

Figure A.3. The main arm body, the points on it, and the projected points $P_{11}$, and $P_{12}$ on it.

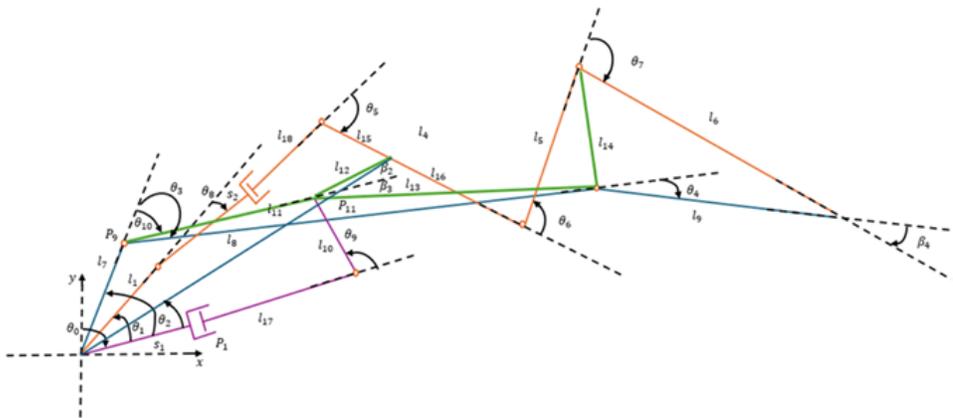

Figure A.4. The kinematic model of the end loader mechanism.

Table A.1. Important parameters on the kinematic model of the end loader mechanism.

| Parameter | Dimension |
|---|---|
| $l_1$ | 348.76 mm |
| $l_2$ | 796.91 mm |
| $l_3$ | 770 mm |
| $l_4$ | 840 mm |
| $l_5$ | 560 mm |
| $l_6$ | 982 mm |
| $l_7$ | 334.23 mm |
| $l_8$ | 2030 mm |
| $l_9$ | 772.33 mm |
| $l_{10}$ | 272.41 mm |
| $l_{11}$ | 1068.88 mm |
| $l_{12}$ | 279.50 mm |
| $l_{13}$ | 973.13 mm |
| $l_{14}$ | 320.88 mm |
| $l_{15}$ | 320 mm |
| $l_{16}$ | 520 mm |
| $\beta_2$ | 26.8° |
| $\beta_3$ | 12.31° |
| $\beta_4$ | 16.06° |
| $\beta_5$ | 5.86° |

Let

$\angle P_i P_j x$: Angle between the vector $P_i P_j$, and x axis

$\angle P_i P_j P_k$: Angle between the vector $P_i P_j$ and $P_i P_k$

$x_p, y_p$: Coordinates of p

The height of the bucket blade ($y_{P8}$) and bucket orientation ($\theta_4$) are utilized in the inverse kinematics to compute the required cylinder extension lengths. $l_i$ and $\beta_i$ represent constant lengths and constant angles, respectively, while $s_i$ and $\theta_i$ denote variable lengths and variable angles. $\beta_0$ and $\beta_1$ are known angles defined as follows:

$$\beta_0 = \pi/2 - \theta_0 + \theta_2 \ (A.1)$$

$$\beta_1 = \pi/2 - \theta_0 + \theta_1 \ (A.2)$$

The variable lengths that depend on the extension of the cylinders are defined as follows:

$$s_{\text{lift}} = s_1 + l_{17} \ (A.3)$$

$$s_{\text{tilt}} = s_2 + l_{18} \ (A.4)$$

The angles $\theta_3$ and $\theta_{10}$ from the model are related through the known angle $\beta_5$ as follows:

$$\theta_3 - \theta_{10} = \beta_5 \ (A.5)$$

The height of the bucket blade ($y_{P8}$) is related to $\theta_3$ through the following equation:

$$y_{P8} = l_7 \sin(\beta_0) + l_8 \sin(\beta_0 - \theta_3) - l_9 \sin(\theta_4 - (\beta_0 - \theta_3)) \ (A.6)$$

Let $\overline{\theta_3} = \beta_0 - \theta_3$:

$$y_{P8} = l_7 \sin(\beta_0) + l_8 \sin(\overline{\theta_3}) - l_9 \sin(\theta_4 - \overline{\theta_3}) \ (A.7)$$

The only unknown in Equation (A.7) is $\overline{\theta_3}$. For the remainder of this section, uppercase letters accompanied by numeric subscripts (e.g., $A_1$) denote constant parameters, whose numerical values can be determined by substituting known values. Equation (A.7) is reformulated into the more compact form shown in Equation (A.8) by introducing the following expressions defined in Equations (A.8) through (A.12):

$$A_1 = y_{P8} - l_7 \sin(\beta_0) \ (A.8)$$

$$A_1 = (l_8 + l_9 \cos(\theta_4)) \sin(\overline{\theta_3})$$

$$- l_9 \sin(\theta_4) \cos(\overline{\theta_3}) \ (A.9)$$

$$B_1 = l_8 + l_9 \cos(\theta_4) \ (A.10)$$

$$C_1 = l_9 \sin(\theta_4) \ (A.11)$$

$$A_1 = B_1 \sin(\overline{\theta_3}) - C_1 \cos(\overline{\theta_3}) \ (A.12)$$

Equation (A.12) can then be rewritten as follows:

$$A_1 = \sqrt{B_1^2 + C_1^2} \sin(\overline{\theta_3} + \phi) \ (A.13)$$

Where the phase angle is expressed as:

$$\phi = \arctan(-C_1/B_1) \ (A.14)$$

$\theta_3$ is then determined as follows:

$$\sin(\overline{\theta_3} + \phi) = A_1/\sqrt{B_1^2 + C_1^2} \ (A.15)$$

$$\overline{\theta_3} + \phi = \arcsin\left(A_1/\sqrt{B_1^2 + C_1^2}\right) \quad (A.16)$$

$$\overline{\theta_3} = \arcsin\left(A_1/\sqrt{B_1^2 + C_1^2}\right) - \phi \quad (A.17)$$

$$\beta_0 - \theta_3 = \arcsin\left(A_1/\sqrt{B_1^2 + C_1^2}\right) - \phi \quad (A.18)$$

$$\theta_3 = \beta_0 - \arcsin\left(A_1/\sqrt{B_1^2 + C_1^2}\right) + \phi \quad (A.19)$$

At this stage, Equation (A.5) is employed to evaluate $\theta_{10}$, which is subsequently utilized to get the coordinates of $P_{12}$ as follows:

$$x_{P12} = l_7 \cos(\beta_0) + l_{11} \cos(\beta_0 - \theta_{10})$$
$$+ l_{12} \cos(\beta_2 + (\beta_0 - \theta_{10})) \quad (A.20)$$

$$y_{P12} = l_7 \sin(\beta_0) + l_{11} \sin(\beta_0 - \theta_{10})$$
$$+ l_{12} \sin(\beta_2 + (\beta_0 - \theta_{10})) \quad (A.21)$$

Utilizing these coordinates, the angle $\angle P_0 P_{12} x$ is calculated as follows:

$$\angle P_0 P_{12} x = \arctan(y_{P12}/x_{P12}) \quad (A.22)$$

$\theta_0$ is obtained from Equation (A.23).

$$(\pi/2 - \theta_0) = \angle P_0 P_{12} x - P_0 P_{12} P_2 \quad (A.23)$$

Equations (A.24) and (A.25) represent the loop closure equations [50,51] for the loop encompassing the links $s_{lift}, l_{10}, l_7$ and $l_{11}$:

$$s_{lift} \cos(\pi/2 - \theta_0) - l_{10} \cos\left(\pi - (\theta_9 + (\pi/2 - \theta_0))\right)$$
$$= l_7 \cos(\beta_0) + l_{11} \cos(\beta_0 - \theta_{10}) \quad (A.24)$$

$$s_{lift} \sin(\pi/2 - \theta_0) + l_{10} \sin\left(\pi - (\theta_9 + (\pi/2 - \theta_0))\right)$$
$$= l_7 \sin(\beta_0) + l_{11} \sin(\beta_0 - \theta_{10}) \quad (A.25)$$

These equations can be expressed as a system of two equations with two unknowns as follows:

$$A_2 x_1 - B_2 \cos(x_2) = C_2 \quad (A.26)$$

$$D_2 x_1 + E_2 \sin(x_2) = F_2 \quad (A.27)$$

Where:

$$x_1 = s_{lift} \quad (A.28)$$

$$x_2 = \pi - (\theta_9 + (\pi/2 - \theta_0)) \quad (A.29)$$

From Equation (A.26):

$$x_1 = \frac{C_2 + B_2 \cos(x_2)}{A_2} \quad (A.30)$$

Rearranging Equation (A.27) yields:

$$D_2 \left(\frac{C_2 + B_2 \cos(x_2)}{A_2}\right) + E_2 \sin(x_2) = F_2 \quad (A.31)$$

$$D_2(C_2 + B_2 \cos(x_2)) + A_2 E_2 \sin(x_2) = F_2 A_2 \quad (A.32)$$

$$D_2 C_2 + D_2 B_2 \cos(x_2) + A_2 E_2 \sin(x_2) = F_2 A_2 \quad (A.33)$$

$$D_2 B_2 \cos(x_2) + A_2 E_2 \sin(x_2) = F_2 A_2 - D_2 C_2 \quad (A.34)$$

Let $G_2 = F_2 A_2 - D_2 C_2$.

$$D_2 B_2 \cos(x_2) + A_2 E_2 \sin(x_2) = G_2 \quad (A.35)$$

Equation (A.35) can be reformulated into the following trigonometric expression:

$$R_2 \sin(x_2 + \phi_2) = G_2 \quad (A.36)$$

Where:

$$R_2 = \sqrt{(A_2 E_2)^2 + (D_2 B_2)^2} \quad (A.37)$$

$$\phi_2 = \arctan(D_2 B_2 / A_2 E_2) \quad (A.38)$$

$\theta_9$ and $s_{lift}$ can be obtained using the following expressions:

$$x_2 + \phi_2 = \arcsin(G_2/R_2) \quad (A.39)$$

$$x_2 = \arcsin(G_2/R_2) - \phi_2 \quad (A.40)$$

$$x_1 = \frac{C_2 + B_2 \cos(x_2)}{A_2} \quad (A.41)$$

$$x_2 = \arcsin\left(\frac{F_2 A_2 - D_2 C_2}{\sqrt{(A_2 E_2)^2 + (D_2 B_2)^2}}\right)$$
$$- \arctan\left(\frac{D_2 B_2}{A_2 E_2}\right) \quad (A.42)$$

$$x_1 =$$
$$\frac{C_2 + B_2 \cos\left(\arcsin\left(\frac{F_2 A_2 - D_2 C_2}{\sqrt{(A_2 E_2)^2 + (D_2 B_2)^2}}\right) - \arctan\left(\frac{D_2 B_2}{A_2 E_2}\right)\right)}{A_2} \quad (A.43)$$

$$s_{lift} = x_1 \quad (A.44)$$

$$\theta_9 = \pi - (\pi/2 - \theta_0) - x_2 \quad (A.45)$$

Utilizing the values obtained for $\theta_0$ and $\theta_3$, the coordinates of $P_7$ and $P_{12}$ can be evaluated as follows:

$$x_{P7} = l_7 \cos(\beta_0) + l_8 \cos(\beta_0 - \theta_3)$$
$$+ l_9 \cos(\theta_4 - (\beta_0 - \theta_3))$$
$$- l_6 \cos\left(\pi/2 - \beta_4 - (\theta_4 - (\beta_0 - \theta_3))\right) \quad (A.46)$$

$$y_{P7} = l_7 \sin(\beta_0) + l_8 \sin(\beta_0 - \theta_3)$$

$$-l_9 \sin(\theta_4 - (\beta_0 - \theta_3)) +$$
$$l_6 \sin\left(\pi/2 - \beta_4 - (\theta_4 - (\beta_0 - \theta_3))\right) \quad (A.47)$$

$$x_{P12} = s_{lift} \cos(\pi/2 - \theta_0)$$
$$-l_{10} \cos\left(\pi - (\theta_9 + (\pi/2 - \theta_0))\right) \quad (A.48)$$

$$y_{P12} = s_{lift} \sin(\pi/2 - \theta_0)$$
$$+l_{10} \sin\left(\pi - (\theta_9 + (\pi/2 - \theta_0))\right) \quad (A.49)$$

The angle $\theta_6$ is obtained using the law of cosines, as follows:

$$l_{\theta 6} = \sqrt{(x_{p7}^2 - x_{p12}^2) + (y_{p7}^2 - y_{p12}^2)} \quad (A.50)$$

$$l_{\theta 6}^2 = l_{16}^2 + l_5^2 - 2l_{16}l_5 \cos(\pi - \theta_6) \quad (A.51)$$

$$\arccos\left(\frac{l_{\theta 6}^2 - l_{16}^2 - l_5^2}{-2l_{16}l_5}\right) = (\pi - \theta_6) \quad (A.52)$$

$$\theta_6 = \left(\pi - \arccos\left(\frac{l_{\theta 6}^2 - l_{16}^2 - l_5^2}{-2l_{16}l_5}\right)\right) \quad (A.53)$$

Utilizing the chain consisting of the links $l_1$, $s_{tilt}$, $l_4$, $l_5$, $l_6$, as well as the chain formed by the links $l_7$, $l_8$, and $l_9$, the following constraint equation can be established:

$$\beta_1 - \theta_8 - \theta_5 + \theta_6 - \theta_7 = \beta_0 - \theta_3 - \theta_4 - \beta_4 \quad (A.54)$$

Equations (A.55) and (A.56) represent the loop closure equations for the loop encompassing the links $s_{lift}$, $l_{10}$, $l_{12}$, $l_{16}$, $l_5$, $l_6$, $l_7$, $l_8$ and $l_9$:

$$s_{lift} \cos(\pi/2 - \theta_0) - l_{10} \cos(\pi - \theta_9 + (\pi/2 - \theta_0))$$
$$+l_{12} \cos(\beta_2 + (\beta_0 - \theta_{10})) + l_{16} \cos(\theta_5 - (\beta_1 - \theta_8))$$
$$+l_5 \cos[\theta_6 - (\theta_5 - (\beta_1 - \theta_8))]$$
$$+l_6 \cos\left[\theta_7 - \left(\theta_6 - (\theta_5 - (\beta_1 - \theta_8))\right)\right]$$
$$= l_7 \cos(\beta_0) + l_8 \cos(\beta_0 - \theta_3)$$
$$+l_9 \cos(\theta_4 - (\beta_0 - \theta_3)) \quad (A.55)$$

$$s_{lift} \sin(\pi/2 - \theta_0) + l_{10} \sin(\pi - \theta_9 + (\pi/2 - \theta_0))$$
$$+l_{12} \sin(\beta_2 + (\beta_0 - \theta_{10})) - l_{16} \sin(\theta_5 - (\beta_1 - \theta_8))$$
$$+l_5 \sin[\theta_6 - (\theta_5 - (\beta_1 - \theta_8))]$$
$$-l_6 \sin\left[\theta_7 - \left(\theta_6 - (\theta_5 - (\beta_1 - \theta_8))\right)\right]$$
$$= l_7 \sin(\beta_0) + l_8 \sin(\beta_0 - \theta_3)$$
$$-l_9 \sin(\theta_4 - (\beta_0 - \theta_3)) \quad (A.56)$$

Isolating $(\theta_5 + \theta_8)$ in Equation (A.54), and then subsequently rearranging Equations (A.55) and (A.56) accordingly, results in the following expressions:

$$s_{lift} \cos(\pi/2 - \theta_0) - l_{10} \cos(\pi - \theta_9 + (\pi/2 - \theta_0))$$
$$+l_{12} \cos(\beta_2 + (\beta_0 - \theta_{10}))$$
$$+l_{16} \cos(\beta_1 + \theta_6 - \theta_7 - (\beta_0 - \theta_3 - \theta_4 - \beta_4) - (\beta_1))$$
$$+l_5 \cos[\theta_6 - (\beta_1 + \theta_6 - \theta_7 - (\beta_0 - \theta_3 - \theta_4 - \beta_4) - (\beta_1))]$$
$$+l_6 \cos\left[\theta_7 - \left(\theta_6 - (\beta_1 + \theta_6 - \theta_7 - (\beta_0 - \theta_3 - \theta_4 - \beta_4) - (\beta_1))\right)\right]$$
$$= l_7 \cos(\beta_0) + l_8 \cos(\beta_0 - \theta_3)$$
$$+l_9 \cos(\theta_4 - (\beta_0 - \theta_3)) \quad (A.57)$$

$$s_{lift} \sin(\pi/2 - \theta_0) + l_{10} \sin(\pi - \theta_9 + (\pi/2 - \theta_0))$$
$$+l_{12} \sin(\beta_2 + (\beta_0 - \theta_{10}))$$
$$-l_{16} \sin(\beta_1 + \theta_6 - \theta_7 - (\beta_0 - \theta_3 - \theta_4 - \beta_4) - (\beta_1))$$
$$+l_5 \sin[\theta_6 - (\beta_1 + \theta_6 - \theta_7 - (\beta_0 - \theta_3 - \theta_4 - \beta_4) - (\beta_1))]$$
$$-l_6 \sin\left[\theta_7 - \left(\theta_6 - (\beta_1 + \theta_6 - \theta_7 - (\beta_0 - \theta_3 - \theta_4 - \beta_4) - (\beta_1))\right)\right]$$
$$= l_7 \sin(\beta_0) + l_8 \sin(\beta_0 - \theta_3)$$
$$-l_9 \sin(\theta_4 - (\beta_0 - \theta_3)) \quad (A.58)$$

The corresponding LCE can then be formulated as the following system of equations:

$$A_3 \cos(x_4) + B_3 \cos(x_4 + \theta_6) = C_3 \quad (A.59)$$
$$D_3 \sin(x_4) + E_3 \sin(x_4 + \theta_6) = F_3 \quad (A.60)$$

Where:
$$A_3 = l_{16} \quad (A.61)$$
$$B_3 = l_5 \quad (A.62)$$
$$D_3 = -l_{16} \quad (A.63)$$
$$E_3 = l_5 \quad (A.64)$$
$$x_4 = (\theta_6 - \theta_7 - (\beta_0 - \theta_3 - \theta_4 - \beta_4)) \quad (A.65)$$

Expanding $B_3 \cos(x_4 + \theta_6)$ and $E_3 \sin(x_4 + \theta_6)$:

$$(A_3 + B_3 \cos(\theta_6)) \cos(x_4) - B_3 \sin(\theta_6) \sin(x_4)$$
$$= C_3 \quad (A.66)$$

$$(D_3 + E_3 \cos(\theta_6)) \sin(x_4) + E_3 \sin(\theta_6) \cos(x_4)$$
$$= F_3 \quad (A.67)$$

The equation system can be reformulated and solved as follows:

$$P = A_3 + B_3 \cos(\theta_6) \quad (A.68)$$

$$Q = -B_3 \sin(\theta_6) \quad (A.69)$$

$$R = D_3 + E_3 \cos(\theta_6) \quad (A.70)$$

$$S = E_3 \sin(\theta_6) \quad (A.71)$$

$$P \cos(x_4) + Q \sin(x_4) = C_3 \quad (A.72)$$

$$R \sin(x_4) + S \cos(x_4) = F_3 \quad (A.73)$$

$$\begin{bmatrix} P & Q \\ S & R \end{bmatrix} \begin{bmatrix} \cos(x_4) \\ \sin(x_4) \end{bmatrix} = \begin{bmatrix} C_3 \\ F_3 \end{bmatrix} \quad (A.74)$$

$$\begin{bmatrix} \cos(x_4) \\ \sin(x_4) \end{bmatrix} = \frac{1}{PR - QS} \begin{bmatrix} R & -Q \\ -S & P \end{bmatrix} \begin{bmatrix} C_3 \\ F_3 \end{bmatrix} \quad (A.75)$$

$$\cos(x_4) = \frac{RC_3 - QF_3}{PR - QS} \quad (A.76)$$

$$\sin(x_4) = \frac{-SC_3 + PF_3}{PR - QS} \quad (A.77)$$

$$x_4 = \operatorname{atan2}(\sin(x_4), \cos(x_4)) \quad (A.78)$$

This solution is used to determine $\theta_7$ from Equation (A.65) as follows:

$$\theta_7 = \theta_6 - x_4 - (\beta_0 - \theta_3 - \theta_4 - \beta_4) \quad (A.79)$$

Using the value of $\theta_7$, the sum $(\theta_8 + \theta_5)$ can be evaluated from Equation (A.54).

Equations (A.80) and (A.81) represent the loop closure equations for the loop encompassing the links $l_1, s_{tilt}, l_{15}, s_{lift}, l_{10}$, and $l_{12}$.

$$l_1 \cos(\beta_1) + s_{tilt} \cos(\beta_1 - \theta_8)$$
$$+ l_{15} \cos(\theta_5 - (\beta_1 - \theta_8))$$
$$= s_{lift} \cos(\pi/2 - \theta_0) - l_{10} \cos(\pi - \theta_9 + (\pi/2 - \theta_0))$$
$$+ l_{12} \cos(\beta_2 + (\beta_0 - \theta_{10})) \quad (A.80)$$

$$l_1 \sin(\beta_1) + s_{tilt} \sin(\beta_1 - \theta_8)$$
$$- l_{15} \sin(\theta_5 - (\beta_1 - \theta_8))$$
$$= s_{lift} \sin(\pi/2 - \theta_0) + l_{10} \sin(\pi - \theta_9 + (\pi/2 - \theta_0))$$
$$+ l_{12} \sin(\beta_2 + (\beta_0 - \theta_{10})) \quad (A.81)$$

These expressions can also be formulated as a system of two equations with two unknowns as follows:

$$x_6 \cos(x_7) = A_4 \quad (A.82)$$

$$x_6 \sin(x_7) = B_4 \quad (A.83)$$

Where:

$$x_6 = s_{tilt} \quad (A.84)$$

$$x_7 = \beta_1 - \theta_8 \quad (A.85)$$

Finally, the resulting system can be solved as follows:

$$(x_6 \cos(x_7))^2 + (x_6 \sin(x_7))^2 = A_4^2 + B_4^2 \quad (A.86)$$

$$x_6^2 (\cos^2(x_7) + \sin^2(x_7)) = A_4^2 + B_4^2 \quad (A.87)$$

$$x_6^2 = A_4^2 + B_4^2 \quad (A.88)$$

$$x_6 = \sqrt{A_4^2 + B_4^2} \quad (A.89)$$

$$\frac{x_6 \sin(x_7)}{x_6 \cos(x_7)} = B_4/A_4 \quad (A.90)$$

$$\tan(x_7) = B_4/A_4 \quad (A.91)$$

$$x_7 = \operatorname{atan2}(B_4, A_4) \quad (A.92)$$

$$\theta_8 = \beta_1 - x_7 \quad (A.93)$$